\documentclass[letterpaper, 10 pt, conference]{ieeeconf}  

\IEEEoverridecommandlockouts                              

\usepackage{graphics} 
\usepackage{xcolor}
\usepackage{hyperref}
\usepackage{epsfig} 
\usepackage{multirow}
\usepackage{subfig}
\usepackage{lipsum}
\usepackage{afterpage}
\usepackage{caption}
\newsavebox{\tempbox}
\title{\LARGE \bf
Reflectivity Is All You Need!: Advancing LiDAR Semantic Segmentation
}

\author{Kasi Viswanath, Peng Jiang and Srikanth Saripalli
\thanks{Kasi Viswanath, Peng Jiang and Srikanth Saripalli are with the Department of Mechanical Engineering, Texas A$\&$M University, College Station, TX-77843
        {\tt\small kasiv,maskjp,ssaripalli@tamu.edu}}%
}

\begin{document}

\maketitle
\thispagestyle{empty}
\pagestyle{empty}

\savebox{\tempbox}{
\begin{minipage}{0.98\textwidth}

\centering
    \includegraphics[width=\textwidth]{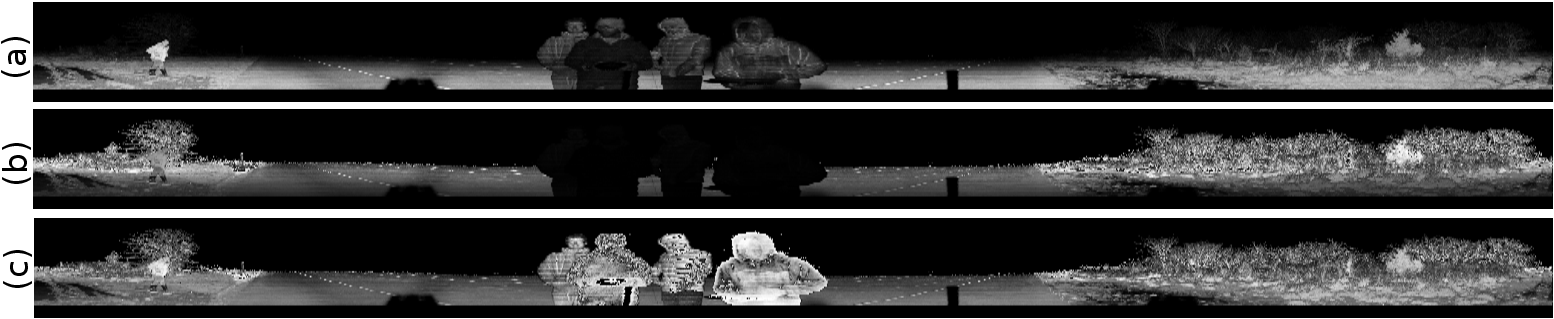}
    \captionof{figure}{Spherical projection of a) Raw intensity data b) Calibrated intensity for Range and angle of incidence($\alpha$) c) Calibrated intensity for near-range ($\eta$), range and angle of incidence($\alpha$).}
    \label{fig:irn_proj}
\end{minipage}}

\begin{figure}[t]
\rlap{\usebox\tempbox}
\vspace{-0.5 cm}
\end{figure}
\afterpage{\begin{figure}[t]
\rule{0pt}{\dimexpr \ht\tempbox+\dp\tempbox}
\vspace{-0.5 cm}
\end{figure}}

\begin{abstract}
LiDAR semantic segmentation frameworks predominantly use geometry-based features to differentiate objects within a scan. Although these methods excel in scenarios with clear boundaries and distinct shapes, their performance declines in environments where boundaries are indistinct, particularly in off-road contexts. To address this issue, recent advances in 3D segmentation algorithms have aimed to leverage raw LiDAR intensity readings to improve prediction precision. However, despite these advances, existing learning-based models face challenges in linking the complex interactions between raw intensity and variables such as distance, incidence angle, material reflectivity, and atmospheric conditions. Building upon our previous work\cite{10.1007/978-3-031-63596-0_54}, this paper explores the advantages of employing calibrated intensity (also referred to as reflectivity) within learning-based LiDAR semantic segmentation frameworks. We start by demonstrating that adding reflectivity as input enhances the LiDAR semantic segmentation model by providing a better data representation. Extensive experimentation with the Rellis-3d off-road dataset shows that replacing intensity with reflectivity results in a 4\% improvement in mean Intersection over Union (mIoU) for off-road scenarios. We demonstrate the potential benefits of using calibrated intensity for semantic segmentation in urban environments (SemanticKITTI) and for cross-sensor domain adaptation. Additionally, we tested the Segment Anything Model (SAM) \cite{kirillov2023segment} using reflectivity as input, resulting in improved segmentation masks for LiDAR images. \textcolor{blue}{\href{https://github.com/unmannedlab/LiDAR-reflectivity-segmentation}{Github repository}}
\end{abstract}


\section{introduction}
Accurate and robust LiDAR semantic segmentation is critical for enabling safe and efficient autonomous navigation, particularly in challenging off-road environments. Existing LiDAR segmentation methods, such as SalsaNext \cite{10.1007/978-3-030-64559-5_16} \cite{9304694}, RangeNet++ \cite{8967762}, Cylinder3D \cite{zhu2020cylindrical}, CENet \cite{cheng2022cenet}, are based primarily on geometric features such as range, position and height. These methods perform well in urban settings with distinct edges and shapes, but often struggle in unstructured off-road scenarios where object shapes are less distinct, making obstacle detection and path planning more difficult.

Recent works such as Point Transformer V3 \cite{wu2024pointtransformerv3simpler}, RandLA-Net \cite{hu2020randlanetefficientsemanticsegmentation} uses novel sampling techniques over geometric features that improve the segmentation of large-scale point clouds but focus on efficiency rather than accuracy. Although these models use LiDAR intensity as an auxiliary input, intensity values are influenced by factors such as range, incidence angle, and surface properties\cite{7337369}, which can distort the data and limit the precision of segmentation.

Reflectivity, a measure of an object's ability to reflect light, offers additional information beyond geometry and presents the potential to improve segmentation accuracy in these complex environments. Unlike raw intensity, calibrated reflectivity is independent of range and angle of incidence, offering a more consistent and interpretable measure of surface characteristics (see Fig. \ref{fig:irn_proj}). By focusing on reflectivity, we can enhance the accuracy of environmental perception in autonomous systems, particularly in off-road settings where traditional intensity data may fall short.

Research on LiDAR intensity has focused primarily on the geospatial domain, particularly in segmenting aerial LiDAR scans to differentiate features such as vegetation, buildings, and roads \cite{rs12101677, s22176388}. In terrestrial laser scanning, calibrated intensity data has been critical for applications like urban roadway lane detection and vegetation monitoring \cite{electronics7110276, 9246255}. As a result, intensity calibration to enhance reflectivity measurement has become a key research objective aimed at improving the interpretation of sensor data. Several innovative approaches have been introduced to address this. For example, \cite{6849466} proposed a novel method for collecting data in controlled environments to estimate parameters for a customized range-intensity equation for terrestrial laser scanners. Similarly, \cite{4432711} introduced an empirical calibration method using brightness targets and a calibrated reference panel under laboratory and field conditions.

\begin{figure}[!t]
    \centering
    \vspace{0.2 cm}
    \includegraphics[width = 0.45\textwidth, height = 5.5 cm]{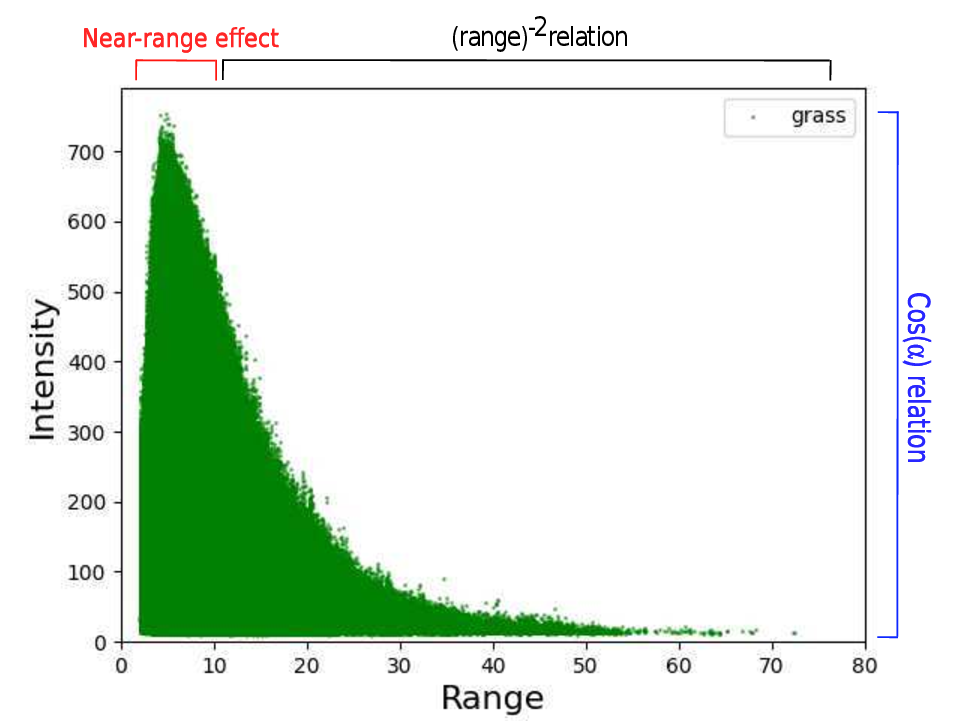}
    \caption{Comparison of raw intensity versus range for the grass class. The figure demonstrates how intensity varies with the range and angle of incidence. It highlights the relationship between $\alpha$ and intensity, indicating that the greatest intensity at each range is achieved when $\alpha$ nears 0, while the lowest intensity occurs when $\alpha$ is approximately $\pi/2$.}
    \label{fig:raw_data}
\vspace{-0.5 cm}
\end{figure}
Recent efforts, including our previous work \cite{10.1007/978-3-031-63596-0_54}, have introduced methods for calibrating intensity values based on range and incidence angle in off-road environments. However, the application was limited to conventional clustering algorithms and a narrow set of object classes, lacking integration with advanced deep learning models. Datasets such as SemanticKITTI \cite{behley2019iccv} and Rellis-3D \cite{9561251} contain raw LiDAR data, yet do not emphasize the potential of reflectivity for enhancing segmentation accuracy.

This paper builds on prior work by exploring how calibrated reflectivity can be combined with geometric features to improve the performance of LiDAR segmentation models. Specifically, we propose a novel approach that fuses reflectivity data with existing neural network architectures to enhance semantic segmentation, particularly in off-road environments. Furthermore, we leverage emerging foundation models such as Segment Anything \cite{kirillov2023segment} to generate segmentation masks using calibrated reflectivity, thus automating part of the annotation process for LiDAR data. The primary contributions of this work are: 
\begin{itemize} 
\item A data-driven approach for calibrating near-range intensity measurements to obtain accurate reflectivity values. 
\item Incorporation of LiDAR reflectivity as an input feature for enhanced data representation in state-of-the-art segmentation models. \item Demonstrating improved segmentation performance in SalsaNext through the integration of reflectivity data. 
\item Using Segment Anything to generate segmentation masks using Range, Intensity, and Reflectivity as input. 
\item A unified LiDAR intensity representation to enable cross-sensor calibration, improving domain adaptation capabilities. \end{itemize}

\section{method}
\begin{figure}
    \centering
    \vspace{0.2 cm}
    \includegraphics[width = 0.4\textwidth]{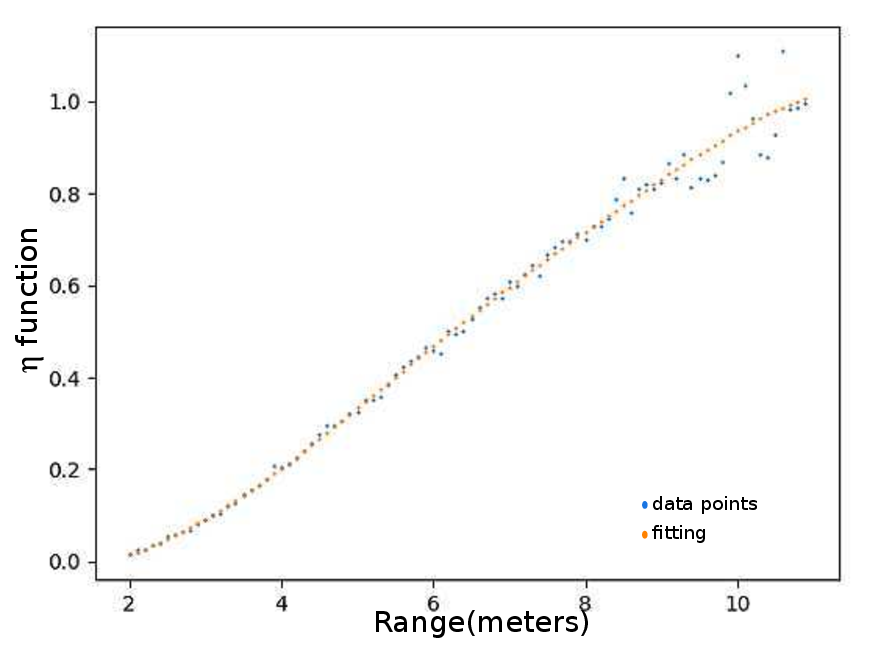}
    \caption{Estimated $\eta(R)$ function for Ouster-64.}
    \label{fig:eta}
\vspace{-0.5 cm}
\end{figure}
\subsection{Revisit LiDAR Intensity Calibration}

\subsubsection{LiDAR Intensity Equation and Near-range Effect} 
The intensity of a LiDAR signal is influenced by the distance to the target (range) and the angle at which the signal strikes the target (angle of incidence). The relationship between these factors is encapsulated by the LiDAR intensity equation \cite{1980easc.conf..546J}:
\begin{equation}
    \centering
    I(R,\alpha,\rho) \propto P_r(R,\alpha,\rho) = \eta(R)\frac{I_e\rho Cos(\alpha)}{R^2}
    \label{ir}
\end{equation}
where $I_e$ denotes the laser's emission power, $R$ signifies the distance to the surface, $\alpha$ represents angle of incidence, and $\rho$ is the parameter indicating the target's reflectivity.

Due to lens defocusing in LiDAR sensors, as detailed by Biavati et.al.\cite{Biavati:11}, a near-range effect, denoted as $\eta(R)$, significantly impacts the intensity data at closer ranges ($R_n < 12$ meters). This phenomenon causes Equation \ref{ir} to be inapplicable within these proximities. We adopt the modeling approach for the near-range effect as outlined in \cite{6849466}, which is described as follows:
\begin{equation}
    \eta(R) = 1 - exp\{\frac{-2r_d^2(R+d)^2}{D^2S^2}\} 
\end{equation}
with $r_d$ denotes the radius of the laser detector, $d$ represents the offset between the measured range and object distance, $D$ is the diameter of the lens, and $S$ signifies focal length, respectively. These parameters can be estimated through controlled experiments \cite{6849466}, which is laborious. We propose a data driven approach to estimate the near-range parameter $\eta(R)$ through semantically annotated datasets.

\subsubsection{Near-range effect parameter estimation}
 For LiDAR points located beyond the influence of the near-range effect, the intensity is solely dependent on the reflectivity coefficient, the range, and the angle of incidence. The reflectivity coefficients for these extended ranges are determined by calibrating the intensity according to the following equation:
\begin{equation}
I_c(\rho) \approx \frac{I_c(R,\alpha,\rho) \times R^2}{\cos(\alpha)} ; R > R_n
\end{equation}
where $R_n$ represents the distance within the influence of near-range effect. Utilizing the annotated labels, the reflectivity coefficients are clustered, yielding a centroid $I_c(\rho)$ for each class. As the reflectivity coefficients remain constant throughout the entire range, $\eta(R)$ function for each class can be derived as follows:
\begin{equation}
\eta(R) = \frac{I_c(R,\alpha,\rho) \times R^2}{I_c(\rho) \times \cos(\alpha)}
\end{equation}

This analysis reveals that the function $\eta(R)$, indicative of the sensor's characteristics, remains consistent across different objects or classes. The function $\eta(R)$ derived for the Ouster sensor within the Rellis-3D dataset is illustrated in Figure \ref{fig:eta}. 
\begin{figure}
\vspace{0.2 cm}
    \centering
    \includegraphics{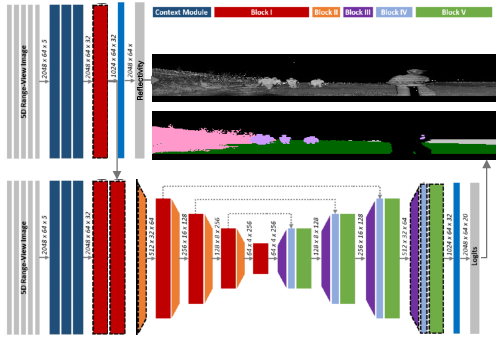}
    \caption{Learning Reflectivity model (Picture modified from SalsaNext\cite{10.1007/978-3-030-64559-5_16})}
    \label{fig:msalsanext}
\vspace{-0.5 cm}
\end{figure}
\subsubsection{Calibrating Intensity}
With $\eta(R)$, range and angle of incidence known, the raw intensity $I(R,\alpha,\rho)$ of a point cloud can be calibrated to provide reflectivity coefficients $I(\rho)$ using the following equation:
\begin{equation}
    I(\rho) = \frac{I(R,\alpha,\rho) \times R^2}{cos(\alpha) \times \eta(R)}
\end{equation}
An example of intensity before and after calibration is shown in Figure \ref{fig:irn_proj}. 
\subsubsection{Cross-Sensor Calibration}
Different LiDAR sensors output intensity data in varying formats. For instance, Velodyne sensors undergo an in-house calibration process for range and power, which is claimed to improve the capabilities of object identification. Given that intensity is a critical feature for semantic segmentation, a model trained on data from one sensor suite often underperforms when applied to data from another sensor, even within identical scenes\cite{DBLP:journals/corr/abs-2103-11351}. This discrepancy poses significant challenges for the robustness of LiDAR segmentation frameworks and impedes effective cross-domain adaptation.

In our previous research \cite{10.1007/978-3-031-63596-0_54}, we introduced a pre-processing technique designed to transform Velodyne data into Ouster format. This transformation is significant because the Ouster format retains the raw form of the LiDAR equation, as illustrated in Figure \ref{fig:raw_data}. The conversion employs a fitting function that is consistently applied across all object classes.
\begin{table*}[!ht]
\vspace{0.3 cm}
    \centering
    \fontsize{7pt}{9pt}\selectfont
    \begin{tabular}{|c|c|c|c|c|c|c|c|c|c|c|c|c|c|c|c|}
         \hline
         Dataset & Config & Grass & Tree & Pole & Vehicle & Log & Person & Fence & Bush & Concrete & Barrier & Puddle & Mud & Rubble & MIoU \\
         \hline
         \multirow{4}{*}{\rotatebox[origin=c]{90}{Rellis-3D}}& rxyzi & 35.1 & 49.6 & \textbf{56.1} & \textbf{23.8} & \textbf{14.5} & 62.8 & 0.6 & 30.2 & 50.0 & 80.1 & 20.4 & 10.7 & 66.6 & 38.4\\ \cline{2-16}
           & rxyzn & \textbf{37.6} & 55.7 & 52.3 & 21.4 & 10.3 & 76.0 & 1.4 & 33.9 & \textbf{74.5} & \textbf{81.3} & \textbf{25.0} & \textbf{13.7} & 67.7 & \textbf{42.4}\\ \cline{2-16}
          & rxyzirn & 36.2 & 55.8  & 51.0 & 20.3 & 9.9 & \textbf{82.7} & 2.7 &  34.0 & 60.1 & 79.1 & 22.6 & 12.9 & \textbf{68.9} & 41.2\\ \cline{2-16}
          & rxyzi$\_$gau & 36.0 & \textbf{58.4} & 52.5 & 22.2 & 12.6 & 82.6 & \textbf{3.2} & \textbf{35.7} & 59.9 & 80.9 & 24.2 & 12.5 & 63.8 & 41.9\\
         \hline
    \end{tabular}
    \caption{mIoU of experiment results. SalsaNext \cite{10.1007/978-3-030-64559-5_16} trained and tested on Rellis-3D with different input configuration. }
    \label{tab:rellis_results}
\end{table*}
\begin{table*}[!ht]
    \centering
    \fontsize{7pt}{9pt}\selectfont
    \begin{tabular}{|c|c|c|c|c|c|c|c|c|c|c|c|c|c|c|c|}
         \hline
         Source & Target & Grass & Tree & Pole & Vehicle & Log & Person & Fence & Bush & Concrete & Barrier & Puddle & Mud & Rubble & MIoU \\
         \hline
         OS rxyzi & VLP rxyzi & 19.6 & \textbf{40.3} & 0.1 & 0.0 & 0.0 & 2.1 & 0.0 & 33.9 & 1.7 & 0.0 & 0.0 & 0.3 & 0.2 & 7.5 \\ 
         \hline
          OS rxyzn & VLP rxyzn & \textbf{30.3} & 38.3 & \textbf{1.1} & \textbf{4.8} & \textbf{0.2} & \textbf{34.1} & \textbf{0.9} & \textbf{35.4} & \textbf{27.3} & \textbf{3.8} & \textbf{1.6} & \textbf{0.9} & \textbf{0.3} & \textbf{13.8}\\ 
         \hline
    \end{tabular}
    \caption{Cross-sensor test on Rellis-3D dataset. OS, VLP refers to Ouster OS-64 and Velodyne VLP-32 respectively.}
    \label{tab:cross_results}
\end{table*}
\begin{table*}[!ht]
    \fontsize{7pt}{11pt}\selectfont
    \resizebox{2.05\columnwidth}{!}{
    \begin{tabular}{|c|c|c|c|c|c|c|c|c|c|c|c|c|c|c|c|c|c|c|c|c|c|}
         \hline
         Dataset & Config & \rotatebox[origin=c]{90}{Car} & \rotatebox[origin=c]{90}{Bicycle} & \rotatebox[origin=c]{90}{Motorcycle} & \rotatebox[origin=c]{90}{Truck} & \rotatebox[origin=c]{90}{Other-vehicle} & \rotatebox[origin=c]{90}{Person} & \rotatebox[origin=c]{90}{Bicyclist} & \rotatebox[origin=c]{90}{Motorcyclist} & \rotatebox[origin=c]{90}{Road} & \rotatebox[origin=c]{90}{Parking} & \rotatebox[origin=c]{90}{Sidewalk} & \rotatebox[origin=c]{90}{Other-ground} & \rotatebox[origin=c]{90}{Building} & \rotatebox[origin=c]{90}{Fence} & \rotatebox[origin=c]{90}{Vegetation} & \rotatebox[origin=c]{90}{Trunk} & \rotatebox[origin=c]{90}{Terrain} & \rotatebox[origin=c]{90}{Pole} & \rotatebox[origin=c]{90}{Traffic Sign} & MIoU \\
         \hline
         \multirow{4}{*}{\rotatebox[origin=c]{90}{SemanticKITTI}} & rxyzi & 90.9 & 37.1 & 51.8 & \textbf{83.4} & 43.0 & 68.7 & 82 & \textbf{0.1} & 77.4 & 44.8 & 60.1 & 6.0 & 73.2 & 45.9 & 46.1 & 64.1 & 47.9 & 56.3 & 46.8 & 54.0\\ \cline{2-22}
           & rxyzn & \textbf{92.4} & \textbf{44.8} & \textbf{52.5} & 71.4 & 46.9  & 69.5 & 84.6 & 0 & 78.6 & 44.4 & 59.6 & 0.3 & 73.6 & \textbf{49.9} & 45.5 & 64.7 & 48.0 & 58.1 & 45.6 & 54.2\\ \cline{2-22}
          & rxyzirn & 92.0 & 41.3 & 50.4 & 75.3 & \textbf{48.5} & \textbf{70.5} & 83.0 & 0 & 77.2 & 43.4 & 59.1 & 0.2 & \textbf{74.9} & 49.8 & \textbf{51.6} & \textbf{66.0} & \textbf{54.0} & 57.0 & \textbf{47.0} & \textbf{54.8} \\ \cline{2-22}
          & rxyzi$\_$gau & 91.9 & 44.4 & 43.6 & 58.6 & 44.7 & 70.2 & \textbf{84.8} & 0.0 & \textbf{79.7} & \textbf{46.5} & \textbf{60.5} & \textbf{10.0} & 73.4 & 48.8 & 45.5 & 64.7 & 48.3 & \textbf{58.6} & 45.3 & 53.7\\
         \hline
    \end{tabular}
    }
    \caption{mIoU of experiment results on SemanticKITTI validation set with different input configuration.  The configuration indicated as $rxyzirn$ achieved the highest mIoU, closely followed by $rxyzn$ and the baseline configuration $rxyzi$ }
    \label{tab:kitti_results}
    \vspace{-0.5 cm}
\end{table*}

\subsection{Leveraging Reflectivity in Deep Learning Models}\ref{sec:md_learn}
To demonstrate the advantages of incorporating reflectivity, we conducted experiments and made modifications to SalsaNext\cite{10.1007/978-3-030-64559-5_16}. Although not the state-of-the-art (SOTA) model on the SemanticKITTI benchmark, SalsaNext excels in off-road scenarios, outperforming the SOTA models on SemanticKITTI. SalsaNext features an encoder-decoder architecture, with the encoder comprising a stack of residual dilated convolutions, and the decoder combining features upsampled from the residual blocks with a pixel-shuffle layer. The architecture is identical to the model shown in Figure \ref{fig:msalsanext}, except that it lacks the reflectivity output head.

\subsubsection{Incorporating Reflectivity as an Input}
The original model processes inputs as five spherical projections \cite{8967762} of the LiDAR scan—$x, y, z$, range, and intensity $[h\times w\times 5]$, as illustrated in Fig. \ref{fig:irn_proj}. It maintains neighborhood features and emphasizes the intensity parameter, effectively treating it as a grayscale image. To substantiate our hypothesis, we replaced the intensity with the calibrated intensity and present our findings in Section \ref{sec:exp_qr}. 

Although range view representations are typically associated with "many-to-one" mapping, semantic inconsistencies, and shape distortions, RangeFormer\cite{10376983} improves accuracy through advanced architectural designs and data-processing techniques, outperforming point, voxel, and multi-view fusion alternatives.

\subsubsection{Learning to Predict Reflectivity}\label{sec:md_learn}
Calibrating intensity for LiDAR data typically requires calculating the surface normal and range for each point, along with using the LiDAR calibration parameters. However, this process can be cumbersome during inference for LiDAR semantic segmentation. To streamline this, we have modified the SalsaNext model by adding an auxiliary head that directly predicts the calibrated intensity. This adaptation allows the model to utilize the standard inputs (i.e., $x, y, z$ coordinates, range, and intensity) without needing additional calibration steps for intensity.

We used an early prediction approach which integrated three convolution layers early in the model to focus on predicting calibrated intensity. Instead of directly using the predicted reflectivity, we took the final feature outputs from this auxiliary head and merged them with the intermediate features from the main SalsaNext model. This approach aims to leverage the calibrated intensity information more effectively within the segmentation process.

 We use calibrated intensity as the ground truth for learning; however, these measurements are not always accurate. To address variability across classes, we model the calibrated intensity for each class using a Gaussian distribution. Rather than employing traditional loss functions such as L1 loss or mean-square loss, we opt for the Gaussian negative log likelihood loss as follows:

\begin{equation}
L=\sum_{c=1}^{classes}\frac{1}{2}\left(\log (var(I_e^c ))+\frac{(I^c-  I_e^c )^2}{var(I_e^c)}\right)
\end{equation}
where $I^c$ represents the input intensity, and $I_e^c$ denotes the calibrated intensity for class $c$.

\section{experimental results}
\subsection{Experiment Setup}
To evaluate the impact of reflectivity on the effectiveness of LiDAR semantic segmentation, we performed multiple experiments.
First, we experiment with how the model will benefit by simply using reflectivity as input. We first trained the SalsaNext model in its original settings, utilizing inputs comprising $x, y, z$ coordinates, $range$, and intensity ($xyzri$). Then, we replace the intensity with two distinct types of reflectivity inputs: 1) calibrated intensity without near-range correction and 2) calibrated intensity with near-range correction. The input configurations were selected as:
\begin{enumerate}
\item range, $x$, $y$, $z$, intensity ($rxyzi$);
\item range, $x$, $y$, $z$, near-range calibrated reflectivity ($rxyzn$); 
\item range, $x$, $y$, $z$, reflectivity, near-range calibrated reflectivity ($rxyzirn$).
\end{enumerate}

In our fourth experimental setup, we assess the performance of the model and loss function introduced in Section \ref{sec:md_learn}, denoting these trials as $rxyzi\_gau$. To maintain consistency across all experiments, key model parameters, including learning rate, weight decay, and batch size, were kept constant, with the exception of the loss function specifically for the $rxyzi\_gau$ configuration. This approach ensures that any variation in performance can be directly attributed to changes in the model and the loss function, rather than external variables.
\begin{figure*}
    \centering
    \vspace{0.2cm}
    \includegraphics[width = 0.98\textwidth ,height = 10 cm]{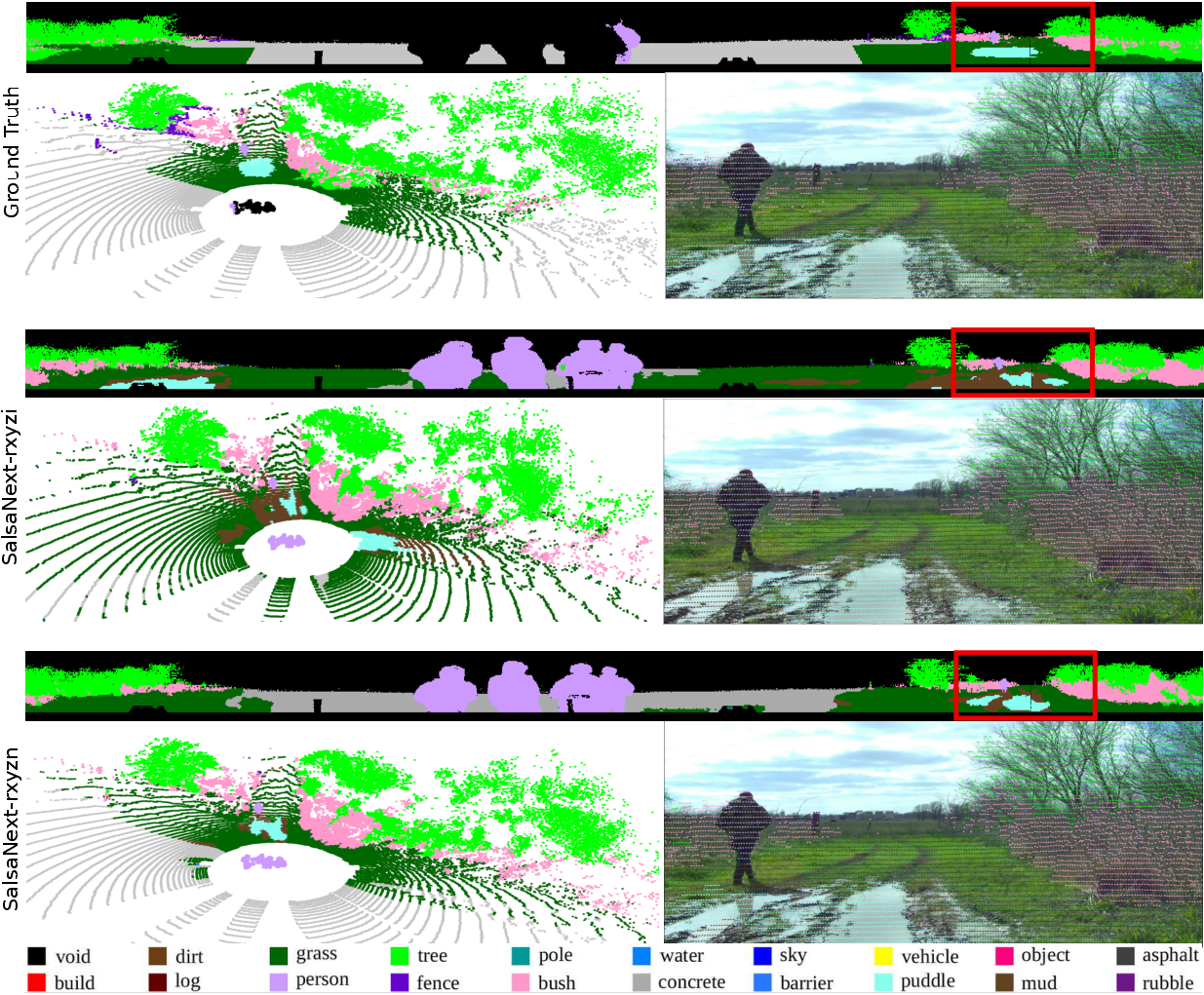}
    \caption{Sample qualitative results from a test set of Rellis-3D. The image shows a spherical projection of point cloud labels for ground truth, Salsanext-$rxyzi$ and Salsanext-$rxyzn$ predictions, along with labeled point cloud projection and camera projections.}
    \label{fig:qual_analysis}
\vspace{-0.5 cm}
\end{figure*}
\subsection{Quantitative Results}\label{sec:exp_qr}
\subsubsection{Off-Road Environment}
Our primary focus is on off-road domains; consequently, we prioritize conducting experiments in off-road environments. For this purpose, we selected the Rellis-3d dataset, which contains semantically annotated LiDAR scans from the Ouster-OS1 64-channel scanner, specifically tailored for off-road scenarios. Performance results are detailed in Table \ref{tab:rellis_results}. When using the $rxyzn$ input configuration, SalsaNext achieved a mean Intersection over Union (mIoU) of $42.4\%$, marking a $4\%$ improvement over the model trained with the $rxyzi$ input. Furthermore, training SalsaNext with the $rxyzirn$ input configuration resulted in a $3\%$ increase in mIoU, underscoring the advantage of incorporating reflectivity for enhanced feature learnability over intensity alone. In another notable result, SalsaNext achieved the second highest mIoU of $41.9\%$ in a fourth setting, illustrating the model's capability to internalize the calibration process, thereby boosting performance. Notably, in this configuration, the model efficiently utilizes only the raw point-cloud data with intensity for inference, showcasing its practical adaptability.
\subsubsection{Cross-sensor}
Rellis-3d also features annotated scans from Velodyne VLP-32 LiDAR for the same environments. It is important to note that Velodyne LiDAR systems output intensity values as precalibrated 8-bit integers. To assess the effectiveness of cross-sensor applicability, we conducted tests in which a SalsaNext model trained on the Ouster $rxyzi$ (intensity) configuration was evaluated on Velodyne $rxyzi$ data and a model trained on Ouster $rxyzn$ (near-range calibrated reflectivity) was tested on Velodyne $rxyzn$. The comparative results, presented in Table \ref{tab:cross_results}, reveal that the model trained on Ouster $rxyzn$ outperformed the one trained on Ouster $rxyzi$, showing a significant $6\%$ increase in mean Intersection over Union (mIoU) and leading in 12 out of 13 categories.

\subsubsection{Urban Environment}
For urban environment testing, we utilize the SemanticKITTI dataset, a multi-modal dataset designed for urban environments, which features annotated LiDAR scans from Velodyne HDL-64E. The performance on SemanticKITTI is detailed in Table \ref{tab:kitti_results}, highlighting the efficacy of different input configurations. The configuration indicated as $rxyzirn$ achieved the highest mIoU at $54.8\%$, followed closely by $rxyzn$ at $54.2\%$, and the baseline configuration $rxyzi$, which also demonstrated competitive performance, achieving a $54.0\%$ mIoU. 
\subsection{Qualitative Results}
Qualitative analysis was performed on Rellis-3d, with SalsaNext input configurations $rxyzi$ and $rxyzn$ as shown in Figure \ref{fig:qual_analysis}. Figure \ref{fig:qual_analysis} compares the predictions with the ground truth represented in spherical projection and camera projection of segmented object points. (Note: SalsaNext doesn't use images for training, and the figure is only for visualization purposes.) The figure shows that the predictions from $rxyzn$ provide better boundaries for puddles and dirt than the predictions of $rxyzi$ and even ground truth. In addition, the ground truth labels the dirt surrounding the puddle region as a puddle, and many more instances of incorrect labeling can be found in the Rellis-3d dataset. This could be attributed to the overall low performance of SalsaNext in all configurations. 

Qualitative analysis of the SemanticKITTI dataset revealed that models using reflectivity and those employing intensity produce similar predictions. This observation underscores the significant contrast or variation in intensity values observed within urban scenes, suggesting that intensity alone can suffice for the effective segmentation of various objects.
\begin{figure*}
    \centering
    \includegraphics[width = 0.98\textwidth, height = 6.5 cm]{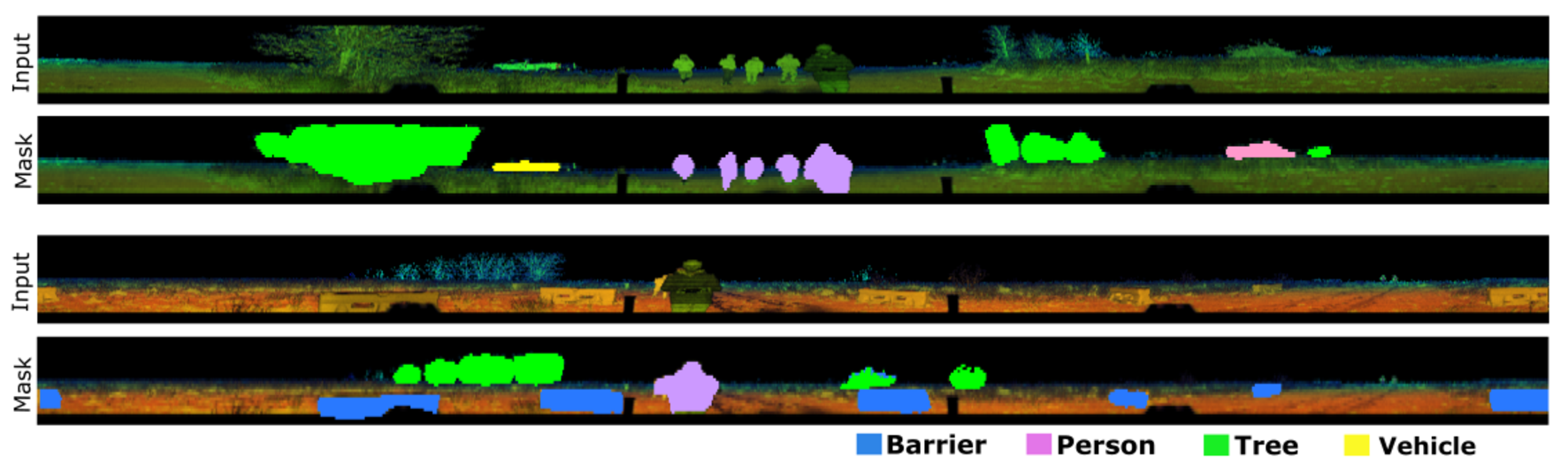}
    \caption{SAM inputs and corresponding annotated masks. The prompts are single point coordinate on the object.}
    \label{fig:SAM_ref}

\end{figure*}
\subsection{Reflectivity on Segment Anything}
In this experiment, we utilized the Segment Anything Model (SAM), a foundation model designed for segmentation via zero-shot generalization using prompts. Specifically, we employed the ViT-H SAM checkpoint, which was trained on over 1 billion segmentation masks across 11 million RGB images. To adapt SAM for LiDAR data, we generated LiDAR images by spherically projecting range, intensity, and reflectivity values, using these as inputs in place of the traditional RGB channels.

As illustrated in Figure \ref{fig:SAM_ref}, SAM produced segmentation masks for the LiDAR inputs, which were subsequently annotated with corresponding labels. The objective of this experiment was to demonstrate the capability of SAM to generate segmentation masks for objects, highlighting the enhanced data representation provided by LiDAR scans through reflectivity. Given SAM’s widespread application in image annotation tasks \cite{cvat}, we present its use case in annotating LiDAR scans, a process that has historically been labor-intensive and time-consuming.

\subsection{Runtime Evaluation}
Given the critical importance of inference speed in the field of autonomous driving, we have meticulously optimized the calibration and prediction processes to facilitate real-time inference. Our framework, which includes LiDAR intensity calibration for near-range, range, and angle of incidence, coupled with SalsaNext prediction, achieves a processing speed of 20 Hz (approximately 50 ms) on an Nvidia RTX 4070 GPU. Considering that LiDAR sensors commonly publish point cloud data at 10 Hz, our methodology ensures real-time inference of scans while maintaining accuracy comparable to other cutting-edge segmentation frameworks.

\section{Inference}
Experiments conducted on the Rellis-3D and SemanticKITTI datasets demonstrate that utilizing reflectivity instead of intensity leads to improved segmentation results in both off-road and on-road scenarios. The significant accuracy boost observed in off-road scenes can be attributed to reflectivity's superior ability to discriminate between classes. This is crucial as objects within a class may share color and texture, while variations exist across classes. The unstructured nature of off-road environments with loose boundaries makes geometry-based features less reliable, leading to lower performance when used alone.

In contrast, in urban settings with well-defined boundaries, geometric features contribute more significantly to learning compared to reflectivity/intensity. Variations in intensity between objects of the same class (e.g., cars with different surface colors) can also affect segmentation. This may explain the limited improvement seen with reflectivity in SemanticKITTI. Nonetheless, incorporating reflectivity consistently improved segmentation compared to intensity across all experiments.

Cross-sensor experiments on Rellis-3D confirm that transferring a model between sensors, even within the same scene, yields poor results. While LiDAR intensity calibration helps, factors like ring count, ray pattern, and field of view significantly impact cross-domain capability.
\section{discussion and conclusion}
In our study, we have shown the promise of using calibrated LiDAR intensity to enhance semantic segmentation. By incorporating reflectivity, Salsanext yielded a 4$\%$ increase in mIoU score on Rellis-3D and exhibited superior boundary predictions compared to intensity alone. Although calibration ensures accuracy, adverse weather conditions such as rain, snow, and fog can disrupt LiDAR performance, resulting in inaccurate point clouds due to particle interference. Therefore, future efforts will focus on resilient sensors such as radar, which can offer precise data even in challenging weather conditions. The task of annotation for images has received wide attention, resulting in significant contribution towards automation algorithms, but pointcloud annotation due to their sparse features remains unexplored. With enhanced data representation of LiDAR, we plan on fine-tuning SAM for LiDAR thereby creating a first-in-series automated annotation tool for pointcloud. Based on these findings, we advocate for the robotics community to transition from intensity to reflectivity and pursue the development of specialized segmentation models that use the reflectivity parameter.





\bibliographystyle{IEEEtran}
\bibliography{IEEEabrv,Reference}

\begin{thebibliography}{10}
\providecommand{\url}[1]{#1}
\csname url@rmstyle\endcsname
\providecommand{\newblock}{\relax}
\providecommand{\bibinfo}[2]{#2}
\providecommand\BIBentrySTDinterwordspacing{\spaceskip=0pt\relax}
\providecommand\BIBentryALTinterwordstretchfactor{4}
\providecommand\BIBentryALTinterwordspacing{\spaceskip=\fontdimen2\font plus
\BIBentryALTinterwordstretchfactor\fontdimen3\font minus \fontdimen4\font\relax}
\providecommand\BIBforeignlanguage[2]{{%
\expandafter\ifx\csname l@#1\endcsname\relax
\typeout{** WARNING: IEEEtran.bst: No hyphenation pattern has been}%
\typeout{** loaded for the language `#1'. Using the pattern for}%
\typeout{** the default language instead.}%
\else
\language=\csname l@#1\endcsname
\fi
#2}}

\bibitem{10.1007/978-3-031-63596-0_54}
K.~Viswanath, P.~Jiang, P.~B. Sujit, and S.~Saripalli, ``Off-road lidar intensity based semantic segmentation,'' in \emph{Experimental Robotics}, M.~H. Ang~Jr and O.~Khatib, Eds.\hskip 1em plus 0.5em minus 0.4em\relax Cham: Springer Nature Switzerland, 2024, pp. 608--617.

\bibitem{kirillov2023segment}
\BIBentryALTinterwordspacing
A.~Kirillov, E.~Mintun, N.~Ravi, H.~Mao, C.~Rolland, L.~Gustafson, T.~Xiao, S.~Whitehead, A.~C. Berg, W.-Y. Lo, P.~Dollár, and R.~Girshick, ``Segment anything,'' 2023. [Online]. Available: \url{https://arxiv.org/abs/2304.02643}
\BIBentrySTDinterwordspacing

\bibitem{10.1007/978-3-030-64559-5_16}
T.~Cortinhal, G.~Tzelepis, and E.~Erdal~Aksoy, ``Salsanext: Fast, uncertainty-aware semantic segmentation of lidar point clouds,'' in \emph{Advances in Visual Computing}, G.~Bebis, Z.~Yin, E.~Kim, J.~Bender, K.~Subr, B.~C. Kwon, J.~Zhao, D.~Kalkofen, and G.~Baciu, Eds.\hskip 1em plus 0.5em minus 0.4em\relax Cham: Springer International Publishing, 2020, pp. 207--222.

\bibitem{9304694}
E.~E. Aksoy, S.~Baci, and S.~Cavdar, ``Salsanet: Fast road and vehicle segmentation in lidar point clouds for autonomous driving,'' in \emph{2020 IEEE Intelligent Vehicles Symposium (IV)}, 2020, pp. 926--932.

\bibitem{8967762}
A.~Milioto, I.~Vizzo, J.~Behley, and C.~Stachniss, ``Rangenet ++: Fast and accurate lidar semantic segmentation,'' in \emph{2019 IEEE/RSJ International Conference on Intelligent Robots and Systems (IROS)}, 2019, pp. 4213--4220.

\bibitem{zhu2020cylindrical}
X.~Zhu, H.~Zhou, T.~Wang, F.~Hong, Y.~Ma, W.~Li, H.~Li, and D.~Lin, ``Cylindrical and asymmetrical 3d convolution networks for lidar segmentation,'' \emph{arXiv preprint arXiv:2011.10033}, 2020.

\bibitem{cheng2022cenet}
H.-X. Cheng, X.-F. Han, and G.-Q. Xiao, ``Cenet: Toward concise and efficient lidar semantic segmentation for autonomous driving,'' in \emph{2022 IEEE International Conference on Multimedia and Expo (ICME)}.\hskip 1em plus 0.5em minus 0.4em\relax IEEE, 2022, pp. 01--06.

\bibitem{wu2024pointtransformerv3simpler}
\BIBentryALTinterwordspacing
X.~Wu, L.~Jiang, P.-S. Wang, Z.~Liu, X.~Liu, Y.~Qiao, W.~Ouyang, T.~He, and H.~Zhao, ``Point transformer v3: Simpler, faster, stronger,'' 2024. [Online]. Available: \url{https://arxiv.org/abs/2312.10035}
\BIBentrySTDinterwordspacing

\bibitem{hu2020randlanetefficientsemanticsegmentation}
\BIBentryALTinterwordspacing
Q.~Hu, B.~Yang, L.~Xie, S.~Rosa, Y.~Guo, Z.~Wang, N.~Trigoni, and A.~Markham, ``Randla-net: Efficient semantic segmentation of large-scale point clouds,'' 2020. [Online]. Available: \url{https://arxiv.org/abs/1911.11236}
\BIBentrySTDinterwordspacing

\bibitem{7337369}
K.~Tan, X.~Cheng, X.~Ding, and Q.~Zhang, ``Intensity data correction for the distance effect in terrestrial laser scanners,'' \emph{IEEE Journal of Selected Topics in Applied Earth Observations and Remote Sensing}, vol.~9, no.~1, pp. 304--312, 2016.

\bibitem{rs12101677}
\BIBentryALTinterwordspacing
A.~Novo, N.~Fariñas-Álvarez, J.~Martínez-Sánchez, H.~González-Jorge, and H.~Lorenzo, ``Automatic processing of aerial lidar data to detect vegetation continuity in the surroundings of roads,'' \emph{Remote Sensing}, vol.~12, no.~10, 2020. [Online]. Available: \url{https://www.mdpi.com/2072-4292/12/10/1677}
\BIBentrySTDinterwordspacing

\bibitem{s22176388}
\BIBentryALTinterwordspacing
W.~Luo, S.~Gan, X.~Yuan, S.~Gao, R.~Bi, and L.~Hu, ``Test and analysis of vegetation coverage in open-pit phosphate mining area around dianchi lake using uav–vdvi,'' \emph{Sensors}, vol.~22, no.~17, 2022. [Online]. Available: \url{https://www.mdpi.com/1424-8220/22/17/6388}
\BIBentrySTDinterwordspacing

\bibitem{electronics7110276}
\BIBentryALTinterwordspacing
J.~Jung and S.-H. Bae, ``Real-time road lane detection in urban areas using lidar data,'' \emph{Electronics}, vol.~7, no.~11, 2018. [Online]. Available: \url{https://www.mdpi.com/2079-9292/7/11/276}
\BIBentrySTDinterwordspacing

\bibitem{9246255}
K.~Tan, W.~Zhang, Z.~Dong, X.~Cheng, and X.~Cheng, ``Leaf and wood separation for individual trees using the intensity and density data of terrestrial laser scanners,'' \emph{IEEE Transactions on Geoscience and Remote Sensing}, vol.~59, no.~8, pp. 7038--7050, 2021.

\bibitem{6849466}
W.~Fang, X.~Huang, F.~Zhang, and D.~Li, ``Intensity correction of terrestrial laser scanning data by estimating laser transmission function,'' \emph{IEEE Transactions on Geoscience and Remote Sensing}, vol.~53, no.~2, pp. 942--951, 2015.

\bibitem{4432711}
S.~Kaasalainen, A.~Kukko, T.~Lindroos, P.~Litkey, H.~Kaartinen, J.~Hyyppa, and E.~Ahokas, ``Brightness measurements and calibration with airborne and terrestrial laser scanners,'' \emph{IEEE Transactions on Geoscience and Remote Sensing}, vol.~46, no.~2, pp. 528--534, 2008.

\bibitem{behley2019iccv}
J.~Behley, M.~Garbade, A.~Milioto, J.~Quenzel, S.~Behnke, C.~Stachniss, and J.~Gall, ``{SemanticKITTI: A Dataset for Semantic Scene Understanding of LiDAR Sequences},'' in \emph{Proc. of the IEEE/CVF International Conf.~on Computer Vision (ICCV)}, 2019.

\bibitem{9561251}
P.~Jiang, P.~Osteen, M.~Wigness, and S.~Saripalli, ``Rellis-3d dataset: Data, benchmarks and analysis,'' in \emph{2021 IEEE International Conference on Robotics and Automation (ICRA)}, 2021, pp. 1110--1116.

\bibitem{1980easc.conf..546J}
A.~V. {Jelalian}, ``{Laser radar systems},'' in \emph{EASCON 1980; Electronics and Aerospace Systems Conference}, Jan. 1980, pp. 546--554.

\bibitem{Biavati:11}
\BIBentryALTinterwordspacing
G.~Biavati, G.~D. Donfrancesco, F.~Cairo, and D.~G. Feist, ``Correction scheme for close-range lidar returns,'' \emph{Appl. Opt.}, vol.~50, no.~30, pp. 5872--5882, Oct 2011. [Online]. Available: \url{https://opg.optica.org/ao/abstract.cfm?URI=ao-50-30-5872}
\BIBentrySTDinterwordspacing

\bibitem{DBLP:journals/corr/abs-2103-11351}
\BIBentryALTinterwordspacing
L.~Wang, D.~Li, Y.~Zhu, L.~Tian, and Y.~Shan, ``Cross-dataset collaborative learning for semantic segmentation,'' \emph{CoRR}, vol. abs/2103.11351, 2021. [Online]. Available: \url{https://arxiv.org/abs/2103.11351}
\BIBentrySTDinterwordspacing

\bibitem{10376983}
L.~Kong, Y.~Liu, R.~Chen, Y.~Ma, X.~Zhu, Y.~Li, Y.~Hou, Y.~Qiao, and Z.~Liu, ``Rethinking range view representation for lidar segmentation,'' in \emph{2023 IEEE/CVF International Conference on Computer Vision (ICCV)}, 2023, pp. 228--240.

\bibitem{cvat}
\BIBentryALTinterwordspacing
``Semantically annotating images using sam.'' [Online]. Available: \url{https://www.cvat.ai/post/facebook-segment-anything-model-in-cvat}
\BIBentrySTDinterwordspacing

\end{thebibliography}

\end{document}